\documentclass[10pt,twocolumn,letterpaper]{article}

\usepackage[pagenumbers]{wacv} 

\usepackage{graphicx}
\usepackage{amsmath}
\usepackage{amssymb}
\usepackage{booktabs}
\usepackage{pgfplots}
\usepackage{hyperref}
\usepackage{tikz}
\usepackage{multirow}
\usepackage{xr}
\usepackage{orcidlink}
\usepackage{pgfplots}
\usepackage{amssymb}
\usepackage{pifont}
\usepgfplotslibrary{fillbetween}
\usepackage[group-separator={,}]{siunitx}
\usepackage{microtype}
\usepackage{xcolor} 
\usepackage{xr}
\usepackage[accsupp]{axessibility}
\usepackage[capitalize]{cleveref}
\crefname{section}{Sec.}{Secs.}
\Crefname{section}{Section}{Sections}
\Crefname{table}{Table}{Tables}
\crefname{table}{Tab.}{Tabs.}

\newcommand*{\addFileDependency}[1]{
  \typeout{(#1)}
  \@addtofilelist{#1}
  \IfFileExists{#1}{}{\typeout{No file #1.}}
}
\makeatother

\definecolor{LIGHTPINK}{RGB}{237,157,202}
\definecolor{LIGHTRED}{RGB}{210,121,121}
\definecolor{LIGHTORANGE}{RGB}{230,170,50}
\definecolor{LIGHTGOLD}{RGB}{210,194,121}
\definecolor{LIGHTGREEN}{RGB}{121,210,121}
\definecolor{LIGHTAQUA}{RGB}{121,206,210}
\definecolor{LIGHTBLUE}{RGB}{121,124,210}
\definecolor{LIGHTPURPLE}{RGB}{153,102,255}
\definecolor{RED}{RGB}{178,34,34}
\definecolor{GRAY}{RGB}{166,166,166}
\definecolor{WHITE}{RGB}{255,255,255}

\definecolor{blue}{rgb}{0.05, 0.50, 0.99}
\definecolor{darkblue}{rgb}{0.0, 0.05, 0.79}
\definecolor{ligthblue}{rgb}{0.0, 0.8, 0.9}
\definecolor{green}{rgb}{0.4, 0.69, 0.2}
\definecolor{darkyellow}{rgb}{0.99, 0.5, 0.01}
\definecolor{yellow}{rgb}{0.9, 0.7, 0.0}
\definecolor{red}{rgb}{0.99, 0.15, 0.07}
\definecolor{darkred}{rgb}{0.8, 0.0, 0.0}

\definecolor{p_orange}{rgb}{0.95, 0.56, 0}
\definecolor{p_purple}{rgb}{0.36, 0.09, 0.40}
\definecolor{p_green}{rgb}{0.25, 0.62, 0.47}
\definecolor{p_blue}{rgb}{0, 0.56, 0.82}
\definecolor{p_red}{rgb}{0.93, 0.24, 0.26}
\definecolor{p_darkblue}{rgb}{0.031, 0.404, 0.533}
\definecolor{p_pink}{rgb}{0.863, 0.498, 0.608}
\definecolor{p_brown}{rgb}{0.298, 0.18, 0.2}

\def\wacvPaperID{1667} 
\def\confName{WACV}
\def\confYear{2025}

\begin{document}

\title{EgoCast: Forecasting Egocentric Human Pose in the Wild}


\author{Maria Escobar \\
Universidad de Los Andes\\
{\tt\small mc.escobar11@uniandes.edu.co}
\and
Juanita Puentes \\
Universidad de Los Andes\\
{\tt\small j.puentes@uniandes.edu.co}
\and
Cristhian Forigua \\
Universidad de Los Andes\\
{\tt\small cd.forigua@uniandes.edu.co}
\and
Jordi Pont-Tuset\\
Google DeepMind\\
{\tt\small jponttuset@google.com}
\and
Kevis-Kokitsi Maninis\\
Google DeepMind\\
{\tt\small kmaninis@google.com}
\and
Pablo Arbelaez\\
Universidad de Los Andes\\
    {\tt\small pa.arbelaez@uniandes.edu.com}
}


\maketitle

\begin{abstract}
Accurately estimating and forecasting human body pose is important for enhancing the user's sense of immersion in Augmented Reality. Addressing this need, our paper introduces EgoCast, a bimodal method for 3D human pose forecasting using egocentric videos and proprioceptive data. We study the task of human pose forecasting in a realistic setting, extending the boundaries of temporal forecasting in dynamic scenes and building on the current framework for current pose estimation in the wild. We introduce a current-frame estimation module that generates pseudo-groundtruth poses for inference, eliminating the need for past groundtruth poses typically required by current methods during forecasting. Our experimental results on the recent Ego-Exo4D and Aria Digital Twin datasets validate EgoCast for real-life motion estimation. On the Ego-Exo4D Body Pose 2024 Challenge, our method significantly outperforms the state-of-the-art approaches, laying the groundwork for future research in human pose estimation and forecasting in unscripted activities with egocentric inputs.
\end{abstract}

\section{Introduction}

As Augmented Reality (AR) continues to revolutionize our interactions within digital environments, accurately capturing the human body's motion becomes essential for delivering a smooth fusion of the real and virtual worlds~\cite{castillo2023bodiffusion, piumsomboon2018mini}.
Beyond capturing the present pose, forecasting human motion is critical for anticipating user actions and needs. Forecasting will be necessary in applications such as patient rehabilitation and medical monitoring~\cite{liu2022lower,natsakis2021predicting,su2023review}, augmented reality responsiveness, and sports coaching~\cite{yuan2019ego}. However,
predicting future motion in real-world settings is challenging because people do not move in simple, predictable patterns. Unlike controlled benchmarks, where movements are designed to reach a specific goal through a limited range of actions, real-life movements are far more varied and complex. Therefore, forecasting methods must be capable of long-term prediction and generalization to different types of actions in the wild. Nonetheless,  forecasting approaches~\cite{martinez2017human, fragkiadaki2015recurrent, jain2016structural} often focus on predefined movements, missing the unpredictable nature of everyday activities. 




\begin{figure*}[t]
    \centering
    \includegraphics[width=0.8\textwidth]{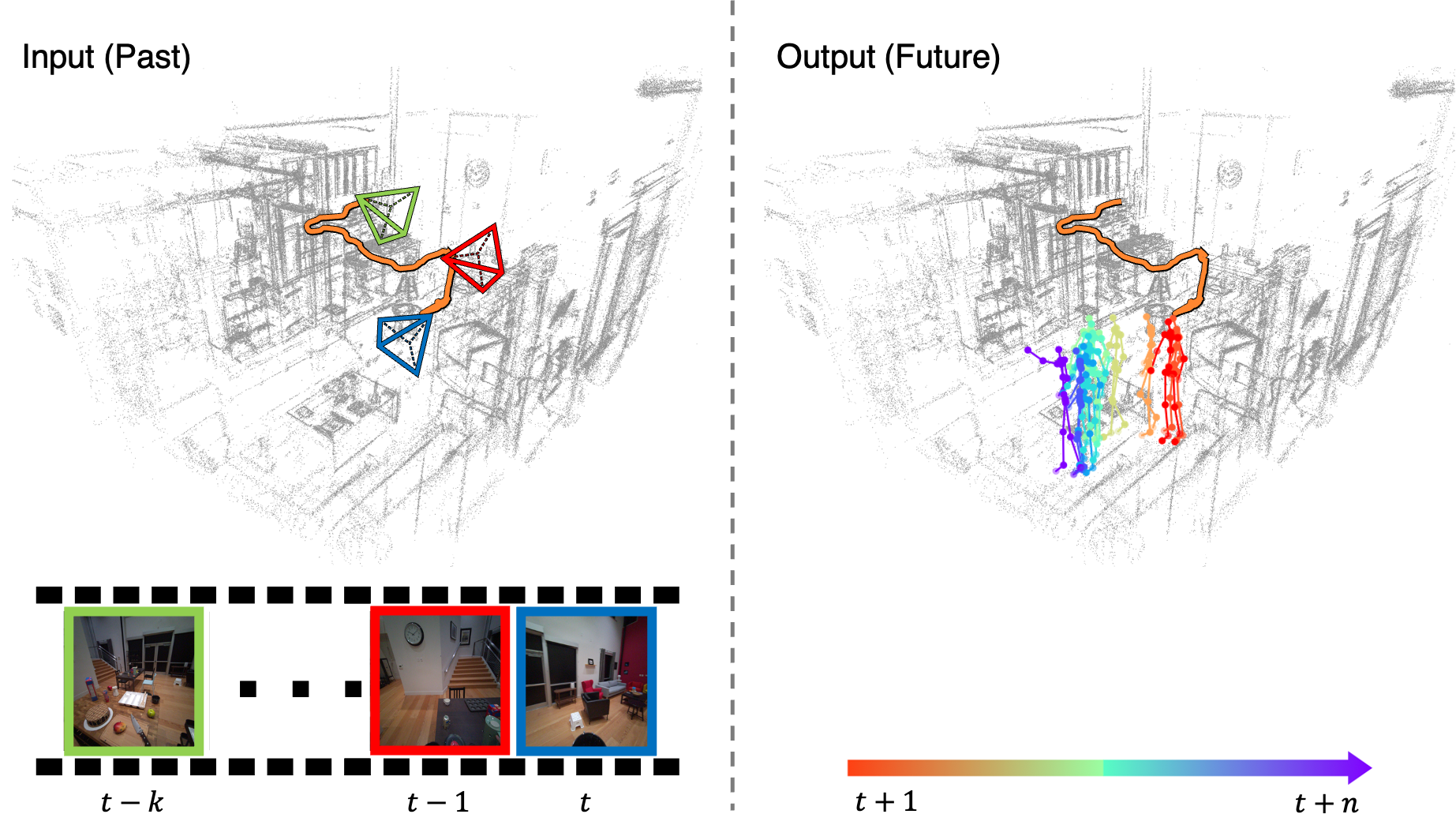}
    \caption{\textbf{3D Human Pose Forecasting.} Our forecasting approach focuses on studying human movement from egocentric inputs in a realistic setting. Given the headset trajectory (3D position and rotation) of the past ($t-k, t$), represented as the orange line in our figure, and the visual cues gathered during the past trajectory, the goal is to forecast the 3D full-body human pose in a future temporal window ($t-t_{n}$), as shown in the right side of the figure. Note that we do not receive as input ground-truth historical poses. }
    \label{fig:task_definition}
\end{figure*}

Besides sight, humans rely on \textit{proprioception} to perceive and manage their body's position and movement.
This internal sensory feedback system is crucial for everyday activities ranging from walking to complex physical motions in fields like gymnastics (\eg{}a gymnast executing a triple jump significantly relies on proprioception to maintain balance and orientation in mid-air).
This proprioceptive awareness is often enhanced by \textit{visual cues}, providing a complete understanding of the surrounding environment and enabling more precise and controlled movements~\cite{luo2022embodied,briscoe2009egocentric}.
This combination of internal sensory feedback and external visual information fits nicely in an egocentric setup that captures a visual stream and head pose as a form of mimicking proprioception.

In this paper, we present EgoCast, a novel method for human pose forecasting in realistic settings, that leverages both internal (proprioceptive) and external (visual) egocentric cues. EgoCast uses as input the 3D proprioceptive information from the head pose along with the visual stream from past observations (Fig.~\ref{fig:task_definition} left) to forecast future 3D human motion (Fig.~\ref{fig:task_definition} right). Our method uses a bimodal Transformer approach that mirrors human perception by mixing proprioceptive and visual information. To avoid relying on ground-truth body poses at test time for forecasting, we first design a current-frame estimation module to predict pseudo-groundtruth full-body poses, given the headset pose and the visual feed from the past. Second, we take the pseudo-groundtruth full-body poses for the past and estimate the future poses using a proprioception encoder that combines the temporal information across frames.

 Current forecasting approaches~\cite{wang2023holoassist, zheng2022gimo} evaluate performance in a short period, usually from 1-5 seconds at 2 FPS, which is not a realistic framework for an AR setting where the user can be constantly moving. In contrast, we propose to assess the forecasting performance in a future span of \{0.5,1,2,3,4,5\} seconds at 30 FPS. Using this setup, we calculate the Mean Per Joint Position Error (MPJPE) at each timeframe and then propose to create an MPJPE curve and use the area under as a new metric to measure the average performance of the methods across all timeframes.


Our results on the egocentric datasets Ego-Exo4D~\cite{grauman2023ego} and Aria Digital Twin (ADT)~\cite{pan2023aria} show that EgoCast is suited for realistic 3D human pose forecasting. Our Current-Frame Estimation Module achieves state-of-the-art performance in the BodyPose challenge of Ego-Exo4D~\cite{grauman2023ego}, outperforming the baseline approach by 4.15 cm and surpassing the previous state-of-the-art by approximately 1 cm, as shown by the results available on the official \href{https://eval.ai/web/challenges/challenge-page/2245/leaderboard/5552}{leaderboard}. Furthermore, our forecasting approach achieves AUC values of 24.41 cm on Ego-Exo4D and 26.69 cm on ADT, showing the benefits of exploiting proprioception alongside visual cues. 

Our main contributions can be summarized as follows:
\begin{itemize}

    \item We formulate a 3D human pose forecasting benchmark for the study of 'in the wild' poses with realistic estimation timeframes and a new evaluation metric. Unlike previous approaches, our forecasting formulation does not rely on ground truth body poses at inference. Instead, we create pseudo-groundtruth poses through our current frame estimation module.  
    
    \item We present EgoCast, a temporally aware transformer-based system that integrates sparse proprioceptive inputs and egocentric visual data for 3D human pose estimation and forecasting. 

\end{itemize}
Find our full project on \href{https://bcv-uniandes.github.io/egocast-wp/}{bcv-uniandes.github.io/egocast-wp/}.

\section{Related Work}

\label{sec:formatting}

\textbf{Egocentric Vision}: AR and robotics applications demand a comprehension of the world from a first-person perspective; thus, egocentric vision has gained significant attention in recent years. Most relevant datasets which collected daily-life activities to study human-object interactions include EPIC-Kitchens~\cite{Damen2022RESCALING, Damen2018EPICKITCHENS, Damen2021PAMI}, UT Ego~\cite{su2016detecting, lee2012discovering}, EGTEA Gaze+~\cite{li2018eye} and Ego4D~\cite{grauman2022ego4d}. However, none of these datasets provides 3D human pose annotations, particularly of the camera wearer. Additional works provide 3D body annotations and visual data; however, their reliance on chest-mounted devices for data collection restricts their practical application in real-world scenarios ~\cite{ng2020you2me, jiang2017seeing}. EgoBody~\cite{zhang2022egobody} introduces a dataset designed for estimating human pose, shape, and motion from first-person perspectives, specifically focusing on social interactions. Nonetheless, the human pose in this dataset is dependent on a specific body model is not suitable for a more generic approach. Furthermore, the Aria Digital Twin (ADT) dataset~\cite{pan2023aria, somasundaram2023project} provides densely annotated 3D human poses and visual data from real-world activities in two realistic environments—an apartment and an office. ADT~\cite{pan2023aria} is specifically designed to focus on realistic long-term activities, serving as a foundational database for developing systems for real-world applications. Recently, Ego-Exo4D~\cite{grauman2023ego} further enriches the landscape of human pose and activity datasets by offering a broad spectrum of human actions captured \textit{in the wild} across diverse environments. 
We leverage Ego-Exo4D~\cite{grauman2023ego} and ADT~\cite{pan2023aria} to examine forecasting in realistic settings.

\textbf{Human Pose Prediction from Sparse Inputs}:
Estimating full-body poses from sparse inputs using head- and hand-tracking devices has become an area of considerable interest within the community. Prior works rely on Inertial Measurements Units (IMUs) to estimate the whole body pose~\cite{yi2022physical,yi2021transpose,huang2018deep}. Huang \textit{et al.} trained a bidirectional LSTM to predict the human body from 6 IMU inputs. Moreover, Jiang \textit{et al.} proposed AvatarPoser~\cite{jiang2022avatarposer}, a transformer-based architecture that integrates traditional Inverse Kinematics (IK) to estimate 3D full-body pose from head-mounted devices and hand controllers. Some recent work~\cite{castillo2023bodiffusion,du2023avatars} leveraged the potential of generative diffusion models for full-body pose estimation. Du \textit{et al.} introduced AGRoL~\cite{du2023avatars}, an MLP-based diffusion model that tracks entire bodies given sparse upper-body tracking signals. 
Recently, Gong \textit{et al.} introduced DiffPose~\cite{gong2023diffpose}, a framework that conceptualizes 3D pose estimation through a reverse diffusion process. Lately, Zhang \textit{et al.} proposed DynaIP, a human pose estimation method using sparse inertial sensors and part-based motion dynamics~\cite{zhang2024dynamic}. However, all these models ignore the external awareness that egocentric visual cues offer.

\textbf{Human Pose Estimation from Visual Inputs:} An alternative line of work is to use egocentric visual cues to estimate 3D full-body human pose. Many existing works utilize downward-facing cameras to capture device users, with several methods employing monocular approaches~\cite{jiang2021egocentric, liu2023egofish3d, tome2019xr, wang2021estimating, akada20243d, wang2024egocentric}. Yuan and Kitani~\cite{yuan2019ego} proposed a reinforcement-learning method for pose estimation that receives egocentric video as input. Kinpoly~\cite{luo2021dynamics} utilizes egocentric video evidence and simulation state to generate a target pose. STCFormer~\cite{tang20233d} models spatio-temporal correlation for 3D human pose estimation. Moreover, Wang \textit{et. al}~\cite{wang2021estimating} introduced a convolutional variational autoencoder-based architecture with a reprojection energy term and a global pose optimizer for pose estimation using a head-mounted fisheye camera. You2Me~\cite{ng2020you2me} predicts the camera wearer's 3D body pose from egocentric video sequences through an LSTM model that leverages the wearer's pose from the preceding frame to predict the pose for the current frame. EgoEgo~\cite{li2023ego} addresses 3D pose estimation in a two-step manner. First, it extracts head motion as an intermediate representation and then uses it for full-body pose prediction. More recently, Wang \textit{et. al}~\cite{wang2023scene} proposed a scene-aware pose estimation network. 
Similarly, xr-EgoPose~\cite{tome2019xr} proposed a convolutional two-step architecture for estimating human pose from synthetic fisheye egocentric videos. Recently,  EgoHMR~\cite{zhang2023probabilistic} estimates 3D human poses from egocentric views by accounting for body truncation, using a diffusion model informed by the surrounding 3D scene. Nevertheless, these methods are entirely based on the external awareness of egocentric images and ignore the critical internal proprioception in human motion.

\begin{figure*}[t]
    \centering
    \includegraphics[width=0.8\textwidth]{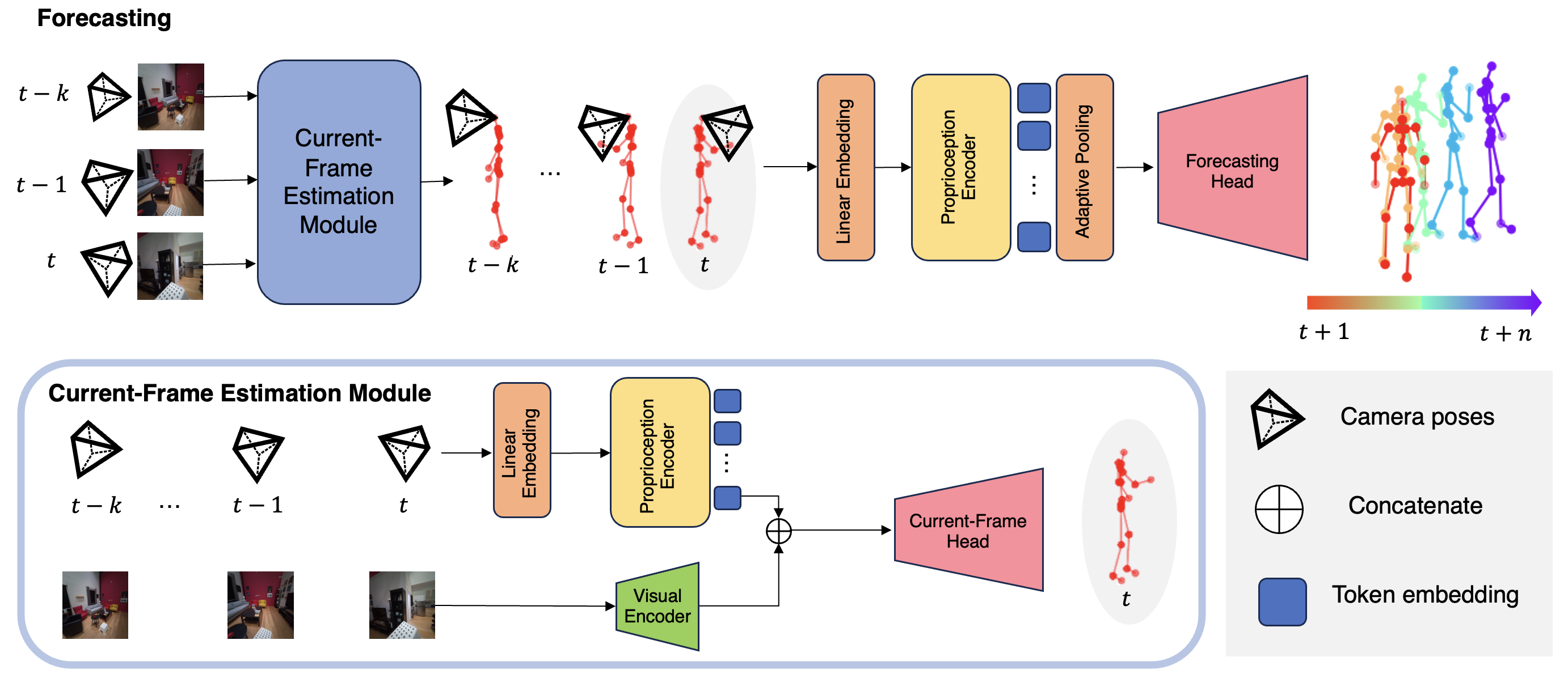}
    \caption{\textbf{EgoCast Overview.}  Our method leverages proprioception and visual streams to estimate 3D human pose. (\textit{Top}) For forecasting, we input previous camera poses and 3D full-body pose predictions through a forecasting head to estimate future 3D poses from $t+1$ to $t+n$. (\textit{Bottom}) Since ground-truth 3D full-body poses are not available in real-case scenarios, we implement a current-frame estimation module that integrates camera poses and visual cues to estimate 3D pose at time $t$.  }
    \label{fig:overview}
\end{figure*}

\textbf{Human Pose Forecasting}: Some works have focused on trajectory forecasting from egocentric inputs~\cite{park2016egocentric, qiu2022egocentric, grauman2022ego4d}, but this line of research only predicts future positions instead of complete human poses. EgoPose~\cite{yuan2019ego} estimates and forecasts 3D human poses from egocentric videos via reinforcement learning. HoloAssist~\cite{wang2023holoassist} is an egocentric human interaction dataset primarily focused on 3D hand forecasting, omitting other body movements. Its predictive model forecasts pose for \num{1.5} seconds at \num{2} FPS from a \num{3}-second input, indicating a need for extended prediction capabilities. Moreover, TEMPO~\cite{choudhury2023tempo} is a framework designed for pose estimation, tracking, and forecasting, leveraging spatiotemporal representations. However, this method is limited to third-person views, and forecasts pose only up to \num{0.33} seconds ahead, or three frames, without considering egocentric perspectives on human motion. Similarly, MotionMixer~\cite{bouazizi2022motionmixer} and SiMLPe~\cite{guo2023back} predict 3D body pose using a multi-layer perceptron model based on previous 3D skeletons, an approach impractical for real-world applications. MotionMixer offers predictions up to \num{4} seconds short-term and \num{10} seconds long-term at \num{2.5} FPS, while SiMLPe extends forecasts up to \num{1} second at \num{25} FPS.
EqMotion~\cite{xu2023eqmotion} introduces a graph-based, efficient equivariant motion prediction model that ensures motion equivariance, capable of predicting pedestrian trajectories up to \num{4.8} seconds at a rate of \num{4} FPS. 
Yan et al. proposed C$^{3}$HOST, a method for forecasting 3D whole-body human poses with a focus on grasping objects~\cite{yan2024forecasting}.  Moreover, GIMO~\cite{zheng2022gimo} integrates eye gaze coordinates, 3D body poses, and contextual scene data through a cross-modal transformer using previous motion patterns. Nonetheless, GIMO's predictive capabilities are limited to forecasting future motion for \num{5} seconds, using an input sequence of \num{3} seconds, at a rate of \num{2} FPS. In contrast, our model, EgoCast, addresses these critical limitations by forecasting human motion over longer timeframes and integrating visual and proprioceptive inputs into the forecasting model.

\section{Human Pose Forecasting Benchmark}

Figure~\ref{fig:overview} shows an overview of our task formulation: predicting a set of 3D human poses in the future given visual and proprioceptive data from the past. We use the term \textit{proprioceptive} to refer to the head pose. EgoCast focuses on a realistic forecasting setting since it does not assume that the models will have access to ground-truth poses from previous frames, and we evaluate on longer than usual timeframes.

\subsection{Task formulation}
At a given point in time $t$, the input is an RGB video sequence  $\mathcal{V}_{RGB} = \{ v^{t-k}, \ldots, v^{t} \}$ and a proprioceptive sequence $\mathcal{P}_{headset} = \{ p^{t-k}, \ldots, p^{t} \} $, $\mathcal{Y}_{headset} = \{ y^{t-k}, \ldots, y^{t} \}$, where $k$ defines the length of the time window in the past, $p^i \in \mathbb{R}^{1\times3}$ denotes the headset position and $y^i \in \phi^{1\times4}$ denotes the headset rotation in quaternion format.
The objective is to predict the sequence of future human poses $\mathcal{Q}_{future} = q^{t+1},\ldots,q^{n}$ where $q^i \in \mathbb{R}^{joints\times3}$ is the 3D position of the joints (17 joints in Ego-Exo4D \cite{grauman2023ego}, 21 joints in ADT~\cite{pan2023aria}) for $n$ future frames.  

\subsection{Evaluation metrics}
We use the Mean Per Joint Position Error (MPJPE) as our primary metric following the standard in pose estimation~\cite{wang2021deep}. The MPJPE measures the average euclidean distance between the ground truth and the predicted 3D joint positions without pre-alignment or root joint trajectory subtraction. We report the MPJPE in cm. We evaluate models for future predictions at \{0.5, 1, 2, 3, 4, and 5\} seconds. Moreover, since the MPJPE evaluates the performance for a fixed future time, we create a curve of the forecasting seconds vs MPJPE and compute the area under the curve (AUC) to measure the overall performance of each method. We perform a \textit{minmax} normalization for AUC computation across time horizons. Since each curve represents errors, the smaller the AUC, the better the model's performance.

\subsection{Datasets}

We use the Ego-Exo4D dataset~\cite{grauman2023ego} as the base of our framework. Ego-Exo4D captures skilled human activities through both egocentric and exocentric perspectives. 
We focus on the egocentric data, which provides insights into close-range hand-object interactions and the wearer's focal points. Egocentric videos are recorded with Aria devices at \num{1404}$\times$\num{1404} resolution and \num{30} FPS.
The activities in Ego-Exo4D~\cite{grauman2023ego} include physical tasks (Soccer, Basketball, Dance and Music) or procedural tasks (Cooking, Bike Repair, and Health) across various locations.
For human pose annotations, Ego-Exo4D~\cite{grauman2023ego} provides a collection of 17 3D joint positions representing the camera wearer's body for each time step. Please refer to Table 1 in the Supplementary Material for additional statistics of Ego-Exo4D~\cite{grauman2023ego}. 

\begin{table*}[h]
\centering

\resizebox{0.75\textwidth}{!}{
\begin{tabular}{@{}c@{\hspace{1em}}c@{\hspace{1em}}c@{\hspace{1em}}c@{\hspace{1em}}c@{\hspace{1em}}c@{\hspace{1em}}c@{\hspace{1em}}c@{\hspace{1em}}|c@{}}
\midrule
\textbf{Method} & \textbf{Basketball} & \textbf{Soccer} & \textbf{Bike repair} & \textbf{Cooking} & \textbf{Health} & \textbf{Dance} & \textbf{Music} & \textbf{Avg.} \\ \hline
EgoEgo~\cite{li2023ego,grauman2023ego}          & 21.36               & 23.08           & 30.18                & 23.71            & 32.57                       & 20.93          & 33.81          & 26,38            \\
Kinpoly~\cite{luo2021dynamics,grauman2023ego}         & 24.98               & 19.09           & 25.19                & 20.80            & 39.23                    & 18.03          & 30.30          & 24.36            \\
Location-based~\cite{castillo2023bodiffusion, jiang2022avatarposer, grauman2023ego}       & 19.89               & 16.62           & 20.61                & 12.65            & 11.63            & 21.15               & 15.00                 & 18.51            \\
EgoCast (Visual-only)   & 16.45           & 17.10        & 13.43               &11.29            & 11.77            & 17.34          & 12.52      & 15.12 

\\
\textbf{EgoCast (Full)} & \textbf{16.31}               & \textbf{14.35}          & \textbf{13.42}               & \textbf{10.24}             & \textbf{10.61}                  & \textbf{16.58}          & \textbf{10.58}           & \textbf{14.36}            \\

\midrule
\end{tabular}}
\caption{\textbf{Ego-Exo4D Current Frame Pose Estimation comparison.} We present the MPJPE for the test split of Ego-Exo4D~\cite{grauman2023ego} in centimeters. EgoCast significantly outperforms the state-of-the-art body pose methods in both individual scenarios and overall performance in~\cite{grauman2023ego}, as shown by the results available on the official \href{https://eval.ai/web/challenges/challenge-page/2245/leaderboard/5552}{leaderboard}.} 
\label{tab:currentClasses}
\end{table*}

Moreover, we utilize the ADT Dataset~\cite{pan2023aria} for comparative analysis. This dataset comprises densely annotated sequences featuring ground-truth 3D body poses by Aria devices~\cite{somasundaram2023project}, including egocentric video recorded at $30$ frames per second, along with the position and rotation of the camera at each frame. Furthermore, ADT~\cite{pan2023aria} includes diverse activities (party, work, decorate, having a meal) showcasing a wide variety of human motions. Although all activities in ADT  take place in the same space, the trajectories vary significantly. We selected ADT for further analysis due to its intentional design for capturing realistic, long-term human activities, which makes it a valuable resource for our research. Please refer to Table 3 in the Supplementary Material for additional statistics of ADT~\cite{pan2023aria}.

\section{EgoCast}

We propose a transformer-based approach for estimating 3D full-body human poses by leveraging two streams of egocentric information: (i) a proprioceptive stream, which includes the headset past positions, and (ii) a visual stream, which includes the RGB egocentric video. 
Both streams are crucial for understanding human motion. 

Figure \ref{fig:overview} shows a schematic overview of our methodology.
Given a sequence of camera poses and RGB images from the headset, we first estimate the 3D full-body pose at each timestamp via the current-frame estimation module.
Then, the pose forecasting module uses these predicted body poses, together with the proprioceptive inputs, to estimate the 3D human poses in the following $n$ frames in the future.
Overall, Egocast proposes a whole pipeline for 3D human pose estimation in real-world scenarios where only the input streams given by the headset are available.

\subsection{Current-Frame Estimation Module}
Given the video sequence 
$ \mathcal{V}_{RGB}$ and proprioceptive sequence $ \mathcal{P}_{headset} $, this module estimates the 3D skeleton, $q^t$, in the current timestamp. 
We take as input tokens the three translation coordinates of the headset from times $t-k$ to $t$. Then, we compute deep embeddings using a linear layer and a transformer-based encoder. We keep the Transformer's output for the last token, corresponding to the encoded proprioceptive information $e^t$. In addition, we also include visual cues for current pose estimation. We use a visual encoder to extract a feature representation $e^v$ from $\mathcal{V}_{RGB}$. Then, compute the final output $q^t$ as 
\begin{equation}
    q^t = \mathcal{H}_c \left( e^t \oplus e^v \right)
\end{equation}
where $ \mathcal{H}_c$ is the current-pose estimation head composed of a 2-layer Multi-Layer Perceptron (MLP). 

At inference time, if fewer frames than the required \( k \) are available, such as at the beginning of the sequence, the module adapts by using only the frames that are present. For example, for the first frame \( t=0 \), the estimation relies solely on the available information from that frame. For the second frame \( t=1 \), the input includes data from frames \( t=0 \) and \( t=1 \). This ensures that the method remains functional and accurate, even when complete temporal context is unavailable in the early stages of a video sequence.

\subsection{Pose Forecasting Module}
Once we have the predicted poses from the current-frame estimation module, we use them as input for the forecasting module. Given past sequences $\mathcal{P}_{headset}$, $\mathcal{Y}_{headset}$ and $\mathcal{Q}_{past}$, we compute input tokens 
\begin{equation}
    \mathcal{T}^j = (q^j  \oplus p^j  \oplus y^j) , t-k\leq j \leq t
\end{equation}
where $\mathcal{T}^j \in \mathbb{R}^{1\times m}$ and $m = (joints \times 3) + 3 + 4$ represents the concatenation of (i) $(joints \times 3)$ for the predicted poses from the current-frame estimation module, (ii) \num{3} from the headset's translation and (iii) \num{4} from the headset's rotation in quaternion format. 
Each timeframe's data is treated as a distinct token within the transformer-based encoder, facilitating the extraction of long-range dependencies inherent in the sequence.
We then extract deep forecasting embeddings $e^f$ by
\begin{equation}
    e^f = \mathcal{E}(l(\mathcal{T}))
\end{equation}
where $e^f\in \mathbb{R}^{k\times d}$, being $d$ the latent dimension of $ \mathcal{E}$, $l$ is a projection layer, and $\mathcal{E}$ is the proprioception encoder that implements a self-attention mechanism to codify each input token and its relations across time and location. Our proprioception encoder includes an adaptive pooling layer to learn a fused representation of all the input tokens. Finally, we pass the fused representation through a 2-layer MLP forecasting head $\mathcal{H}_f$ to compute the final output $\mathcal{Q}_{future}$ as well as a prediction for the translation $\mathcal{P}_{future}$ and rotation of the headset $\mathcal{Y}_{future}$. 

\begin{figure*}[t]
    \centering
    \includegraphics[width=0.8\textwidth]{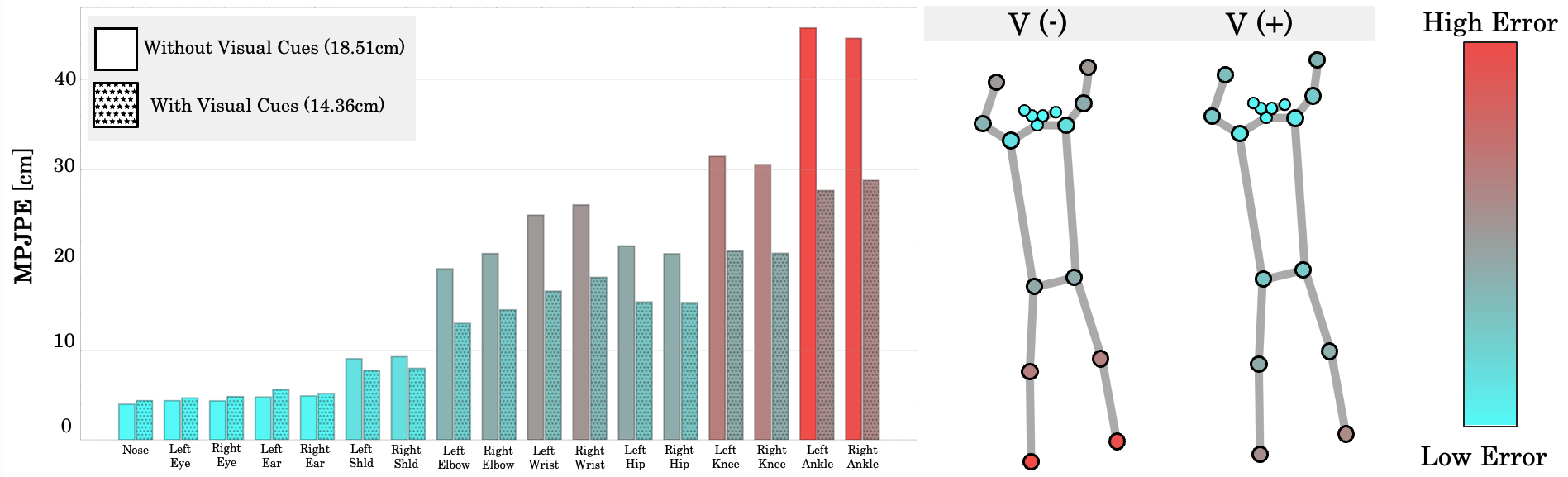}
    \caption{\textbf{Effect of visual cues on MPJPE for the current-frame estimation module.}  For each joint, we present the Mean Per-Joint Position Error (MPJPE) variation, contrasting conditions without visual cues against those with visual cues, through a color scale from blue (low error) to red (high error). Visual egocentric data significantly reduces errors, especially in the lower body.}
    \label{fig:jointsColors}
\end{figure*}
\begin{figure*}[h]
    \centering
    \includegraphics[width=0.85\textwidth]{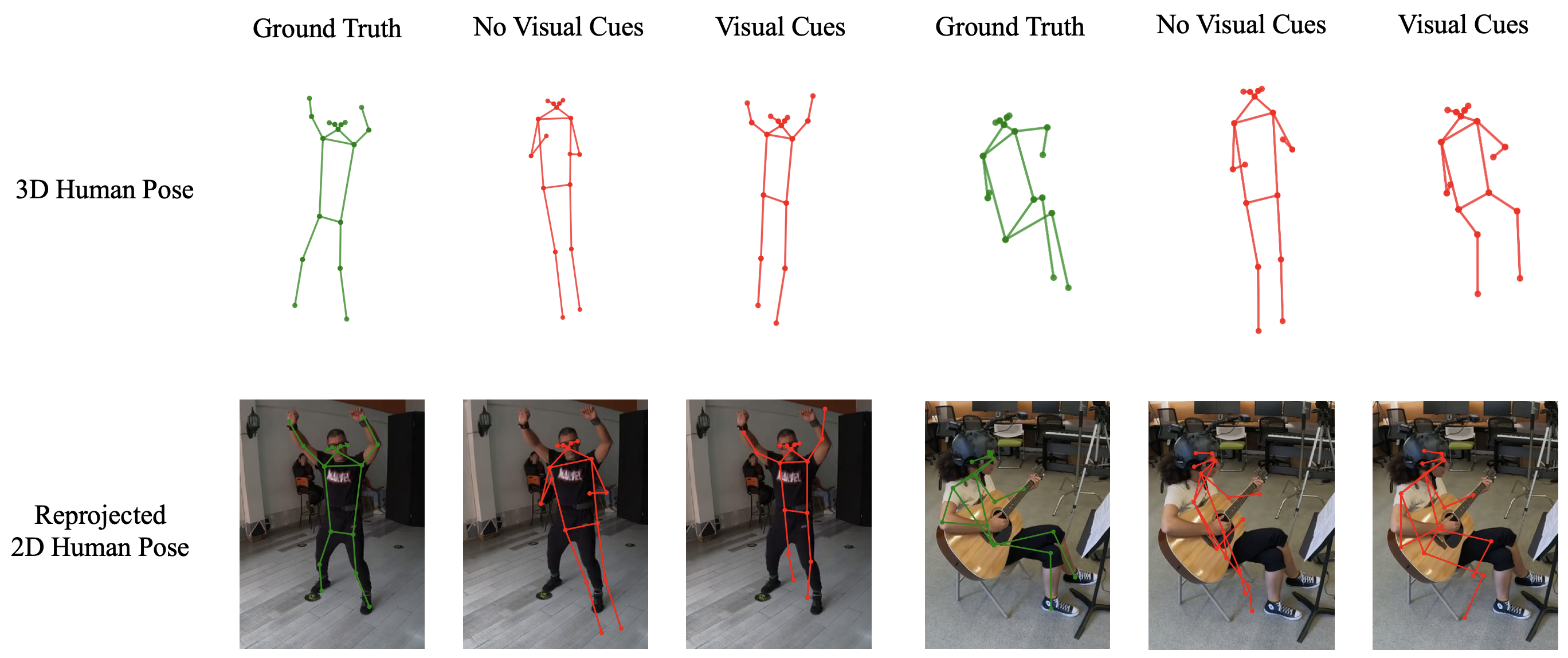}
    \caption{\textbf{3D Human Pose Estimation with and without Visual Inputs.} A common assumption in human pose estimation is that individuals always stand with their hands by their sides. However, integrating visual information into our Current-Frame Estimation module challenges this notion, accurately predicting when a person sits down or raises their hands.
    }
    \label{fig:skels}
\end{figure*}

For training, we compute the $L_1$ distance between the ground truth and predicted 3D full-body poses, headset translations, and rotations in the future as:
\begin{equation}
\label{eq:loss}
    L_f = \lambda_\mathcal{Q}L_1\left(\mathcal{Q},\hat{\mathcal{Q}}\right) + \lambda_RL_1\left(\mathcal{Y},\hat{\mathcal{Y}}\right) +\lambda_TL_1\left(\mathcal{P},\hat{\mathcal{P}}\right)
\end{equation}
where $\mathcal{Q}, \mathcal{R}$, and $\mathcal{P}$ correspond to the ground-truth sequences and $\hat{\mathcal{Q}}, \hat{\mathcal{Y}}$ and $\hat{\mathcal{P}}$ are the predictions for 3D full-body poses and rotation and translation of the headset.

\section{Experiments} 
\label{sec:formatting}
We perform extensive experimentation to assess the performance of our proposed method. In Section \ref{sec:current}, we detail the performance outcomes of the Current-Frame Estimation Module in estimating the 3D skeleton. Subsequently, Section \ref{sec:forecasting} presents our results on forecasting the future sequence of human poses, using the predicted pseudo-groundtruth poses created in the Current-Frame Estimation Module.

\subsection{Current-Frame Pose Estimation}\label{sec:current}

Table \ref{tab:currentClasses} presents the comparative evaluation results for the state-of-the-art methods detailed in Grauman \textit{et al}.~\cite{grauman2023ego} from the Ego-Exo4D~\cite{grauman2023ego} EgoPose task. We also compare our performance when using only visual streams as input (Visual-only) versus mixing proprioceptive and visual streams (Full). EgoCast significantly improves over all other state-of-the-art approaches, reducing the error by 22\%. As shown in Table \ref{tab:currentClasses}, EgoCast performs better in all evaluated scenarios, significantly reducing the MPJPE. Current state-of-the-art approaches utilize only the proprioceptive information (Location-based) or have sequential stages for using the proprioceptive and visual information (EgoEgo and Kinpoly). Using only visual inputs for EgoCast results in an improvement of 4.15 cm in comparison to the best-performing method in Ego-Exo4D~\cite{grauman2023ego}. Furthermore, the additional improvement of EgoCast (Full) shows that merging both information streams since the beginning is a strategy for exploiting the interaction between the internal and external cues for human pose estimation.

Notably, EgoCast achieves significant MPJPE reduction in the \textit{Music} and \textit{Health} scenarios. This can be attributed to the conventional assumption that subjects are standing; however, incorporating visual cues through EgoCast enables the accurate recognition of a seated posture, particularly affecting performance in the lower body joints (Figure~\ref{fig:skels}). An in-depth view of this enhancement is shown in Figure~\ref{fig:jointsColors}, which represents the average error for each joint across all test sequences, using a spectrum from blue (low error) to red (high error). While the accuracy for facial key points (eyes, ears, nose) remains essentially the same with the addition of visual cues, there is a notable improvement in the accuracy for extremities. Through visual data integration, EgoCast improves the accuracy of lower-body joints, notably in the knees and ankles. Remarkably, the error in the ankles is reduced by 39\%, emphasizing EgoCast's capability in accurately predicting the position of lower-body joints, which is challenging without the context provided by visual cues. Furthermore, the \textit{Soccer} scenario sees the most significant improvement when transitioning from the visual-only approach to the full approach. This improvement arises because soccer involves extensive movement across the playing area, making the proprioceptive input highly relevant for capturing precise motion dynamics. The analysis of activities requiring rapid upper body movements, such as dance and basketball, shows that integrating visual data significantly reduces inaccuracies in predicting arm joint positions. Figure~\ref{fig:skels} demonstrates that, without visual information, the model inaccurately predicts the individual's arms as lowered instead of raised. In contrast, incorporating egocentric visual inputs leads to accurate predictions of the arms in a raised position. Figure~\ref{fig:jointsColors} shows this enhancement, especially in the shoulders, elbows, and wrists, highlighted by blue tones, thanks to the addition of visual information.


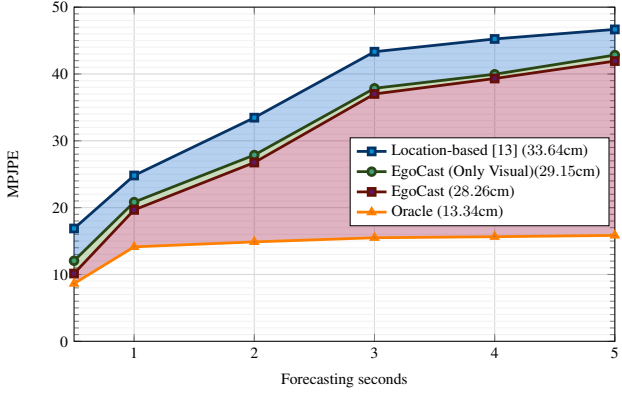
\begin{figure}
\centering
\begin{subfigure}[t]{\linewidth} 
\resizebox{\linewidth}{!}{%
  \begin{tikzpicture}[/pgfplots/width=1.5\linewidth, /pgfplots/height=1\linewidth]
    \begin{axis}[
                 ymin=0,ymax=50,xmin=0.5,xmax=5,
        		 xlabel=Forecasting seconds,
        		 ylabel=MPJPE,
         		 xlabel shift={-2pt},
        		 ylabel shift={-3pt},
		         font=\small,
		         axis equal image=false,
		         enlargelimits=false,
		         clip=true,
        	     grid style=solid, grid=both,
                 major grid style={white!85!black},
        		 minor grid style={white!95!black},
                 ytick={0,10,...,50},
                 yticklabels={0,10,20,30,40,50},
		 		 xtick={0,1,...,5},
                 xticklabels={0.5,1,2,3,4,5},
         		 minor xtick={0,1,...,5},
		         minor ytick={0,1,...,50},
        		 legend style={at={(0.51,0.61)},
                    anchor=north west, font=\small},
                legend cell align={left}]

    \addplot+[name path = baseline, black!60!blue,solid,mark options={solid,fill=p_blue,scale=1},mark=square*,ultra thick] table[x=Secs,y=EgoCast_novisual]{images/main_results.txt};
    \addlegendentry{Location-based~\cite{grauman2023ego} (\num{33.64}{cm})}
    \addplot+[name path = visual, black!60!green,solid,mark=*,mark options={fill=p_green, scale=1.1},ultra thick] table[x=Secs,y=EgoCast_onlyvisual]{images/main_results.txt};
    \addlegendentry{EgoCast (Only Visual)(\num{29.15}{cm})}
    \addplot+[name path = egocast, black!60!red,solid,mark options={solid,fill=p_purple,scale=1},mark=square*,ultra thick] table[x=Secs,y=EgoCast]{images/main_results.txt};
    \addlegendentry{EgoCast (\num{28.26}{cm})}
    \addplot+[name path = oracle, orange,solid,mark options={solid,fill=p_purple,scale=1},mark=triangle*,ultra thick] table[x=Secs,y=Upper]{images/main_results.txt};
    \addlegendentry{Oracle (\num{13.34}{cm})}

    \addplot[black!20!purple, opacity=0.3] fill between [of=egocast and oracle];
    \addplot[black!20!green, opacity=0.3] fill between [of=egocast and visual];
    \addplot[black!20!blue, opacity=0.3] fill between [of=visual and baseline];

    \end{axis}
\end{tikzpicture}
}
\end{subfigure}
\caption{\textbf{Ego-Exo4D Forecasting at different timeframes.} We show performance curves for forecasting at \{0.5, 1, 2, 3, 4, and 5\} seconds in the future. We compare our final EgoCast approach against a forecasting extension of the current state-of-the-art method for current-frame pose estimation and an Oracle approach that aligns the trajectories with the ground truth. Note that since the graph shows MPJPE, lower curves represent better performance. }
\label{fig:egoexo_forecasting}
\end{figure}
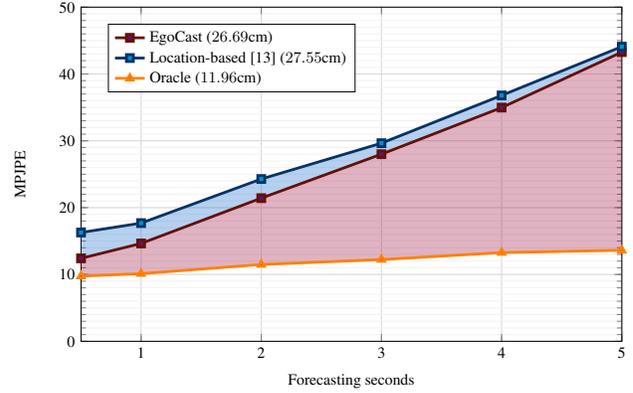
\begin{figure}
   \centering
   \begin{subfigure}[t]{\linewidth} 
\resizebox{\linewidth}{!}{
   \begin{tikzpicture}
   [/pgfplots/width=1.5\textwidth, /pgfplots/height=1\textwidth]
      \begin{axis}[
                   ymin=0,ymax=50,xmin=0.5,xmax=5,
                   xlabel=Forecasting seconds,
                   ylabel=MPJPE,
                   xlabel shift={-2pt},
                   ylabel shift={-3pt},
                   font=\small,
                   axis equal image=false,
                   enlargelimits=false,
                   clip=true,
                   grid style=solid, grid=both,
                   major grid style={white!85!black},
                   minor grid style={white!95!black},
                   ytick={0,10,...,50},
                   yticklabels={0,10,20,30,40,50},
                   xtick={0,1,...,5},
                   xticklabels={0.5,1,2,3,4,5},
                   minor xtick={0,1,...,5},
                   minor ytick={0,1,...,50},
                   legend style={at={(0.05,0.95)},
                                 anchor=north west, font=\small},
                   legend cell align={left}]
                   
      \addplot+[name path = egocast, black!60!red,solid,mark options={solid,fill=p_purple,scale=1},mark=square*,ultra thick] table[x=Secs,y=Ours(pred)]{images/skeleton_ablation.txt};
      \addlegendentry{EgoCast (\num{26.69}{cm})}
      \addplot+[name path =imu, black!60!blue,solid,mark options={solid,fill=p_blue,scale=1},mark=square*,ultra thick] table[x=Secs,y=norot]{images/skeleton_ablation.txt};
      \addlegendentry{Location-based~\cite{grauman2023ego} (\num{27.55}{cm})}
      \addplot+[name path = oracle, orange,mark options={solid,fill=p_purple,scale=1},mark=triangle*,ultra thick] table[x=Secs,y=Upperbound]{images/figure1.txt};
      \addlegendentry{Oracle (\num{11.96}{cm})}

  
    \addplot[black!20!purple, opacity=0.3] fill between [of=egocast and oracle];
    \addplot[black!20!blue, opacity=0.3] fill between [of=egocast and imu];

      \end{axis}
   \end{tikzpicture}}
   \end{subfigure}
   \caption{\textbf{ADT Overall Forecasting performance.} We compare our final EgoCast approach against a forecasting extension of the current state-of-the-art method for current-frame pose estimation and an Oracle approach that aligns the trajectories with the ground truth.}
   \label{fig:ADT_results}
\end{figure}

\begin{figure*}[h]
    \centering
    \includegraphics[width=0.8\textwidth]{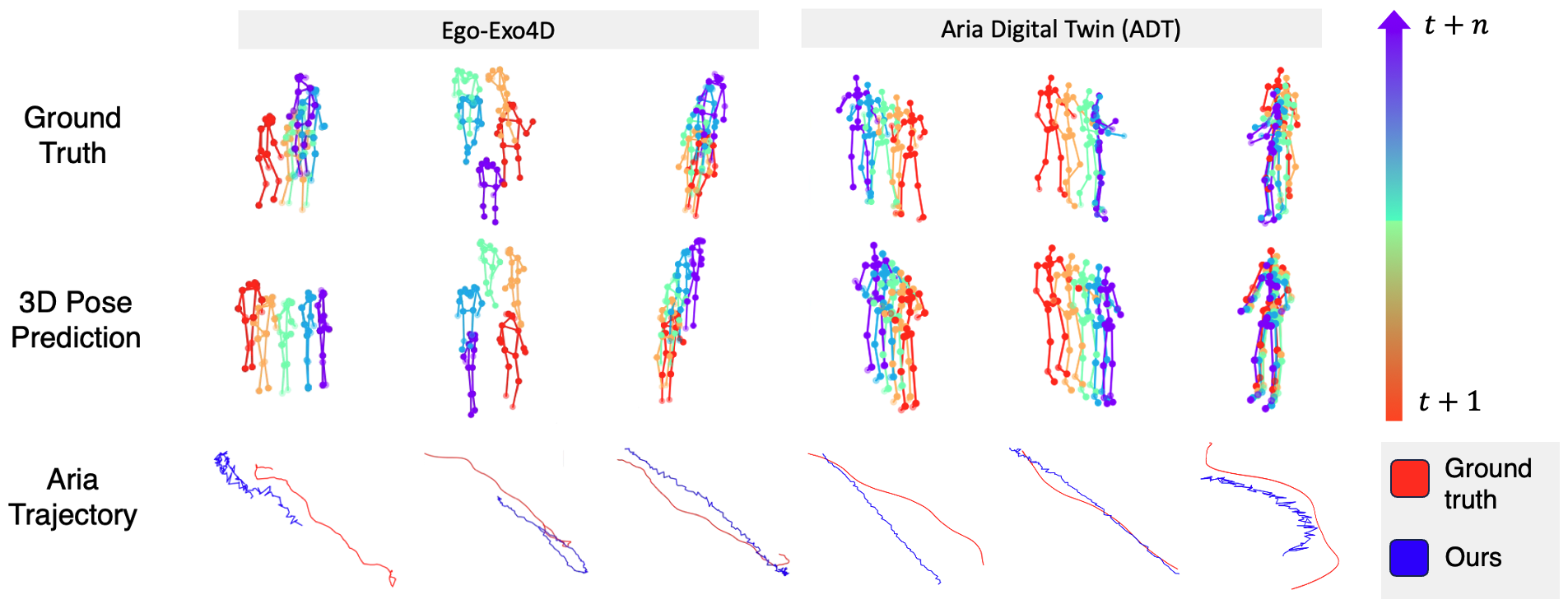}
    \caption{\textbf{Forecasting qualitative results}. We compare qualitatively the predicted 3D human poses against the ground-truth sequences. We also show a comparison of the trajectory prediction. We use an input window size of 20 and forecast 4 seconds (120 frames). Note that we subsampled the results for visualization. Our results show a high resemblance of the 3D poses and realistic trajectory estimation. }
    \label{fig:qualitative}
\end{figure*}

\begin{figure}[t]
   \centering
  \resizebox{\linewidth}{!}{%
  \begin{tikzpicture}[/pgfplots/width=2.6\linewidth, /pgfplots/height=1.63\linewidth]
    \begin{axis}[
                 ymin=0,ymax=140,xmin=0.5,xmax=5,
        		 xlabel=Forecasting seconds,
        		 ylabel=MPJPE,
         		 xlabel shift={-2pt},
        		 ylabel shift={-3pt},
		         font=\large,
		         axis equal image=false,
		         enlargelimits=false,
		         clip=true,
        	     grid style=solid, grid=both,
                 major grid style={white!85!black},
        		 minor grid style={white!95!black},
        		 ytick={0,20,...,140},
                 yticklabels={0,20,40,60,80,100,120,140},
                 xtick={0,1,...,5},
                 xticklabels={0.5,1,2,3,4,5},
         		 minor xtick={0,1,...,5},
		         minor ytick={0,1,...,140},
        		 legend style={at={(0.03,0.97)},
                 		       anchor=north west, font=\Large,},
                 legend cell align={left}, label style={font=\LARGE}]
    \addplot+[p_orange,solid,mark=*,mark options={fill=p_orange},ultra thick] table[x=secs,y=soccer]{images/egoexo.txt};
    \addlegendentry{Soccer (\num{83.68}{cm})}
    
    \addplot+[p_purple,solid,mark=*,mark options={fill=p_purple},ultra thick] table[x=secs,y=basketball]{images/egoexo.txt};
    \addlegendentry{Basketball (\num{84.20}{cm})}

    \addplot+[p_brown,solid,mark=*,mark options={fill=p_brown},ultra thick] table[x=secs,y=cooking]{images/egoexo.txt};
    \addlegendentry{Cooking (\num{25.19}{cm})}
    
    \addplot+[p_green,solid,mark=*,mark options={fill=p_green},ultra thick] table[x=secs,y=dance]{images/egoexo.txt};
    \addlegendentry{Dance (\num{35.54}{cm})}
    
    \addplot+[p_pink,solid,mark=*,mark options={fill=p_pink},ultra thick] table[x=secs,y=bike]{images/egoexo.txt};
    \addlegendentry{Bike (\num{41.44}{cm})}
    
    \addplot+[p_blue,solid,mark=*,mark options={fill=p_blue},ultra thick] table[x=secs,y=music]{images/egoexo.txt};
    \addlegendentry{Music (\num{9.02}{cm})}
    
    \addplot+[p_red,solid,mark=*,mark options={fill=p_red},ultra thick] table[x=secs,y=health]{images/egoexo.txt};
    \addlegendentry{Health (\num{11.49}{cm})}

    \end{axis}

\end{tikzpicture}}
   \caption{\textbf{Forecasting performance by activity.} For each activity in the Ego-Exo4D dataset, we show the performance of our method when forecasting up to 5 seconds into the future. Note that since the graph shows MPJPE, lower curves represent better performance. }
  \label{fig:egoexo_categories}
 \end{figure}
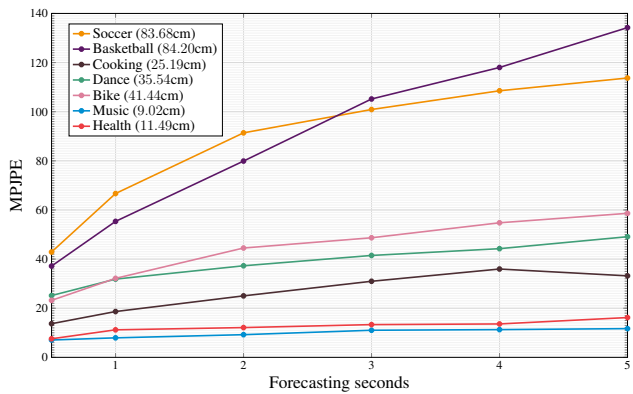

Furthermore, Table 4 in the Supplementary Material presents the ablation performance of the current-frame estimation module on ADT~\cite{pan2023aria}, examining the use of visual streams and variations in window size. Similar to the findings of EgoCast on EgoExo-4D~\cite{grauman2023ego}, the best improvement occurs when visual streams are integrated, highlighting the crucial role of combining proprioceptive inputs with visual data to achieve greater accuracy in pose estimation.

\subsection{Pose Forecasting }\label{sec:forecasting}
Once we have the predictions from the current-frame estimation module, we use them as inputs for our forecasting module. We extend the previous best-performing method of Ego-Exo4D (Location-based in Tab. \ref{tab:currentClasses}) to forecasting for a comprehensive state-of-the-art comparison. Furthermore, we establish an Oracle approach to understand the limitations of our method. Our Oracle is designed to align the translation of predicted trajectories with the ground truth, focusing primarily on the precision of the predicted 3D poses. 

As shown in Figure \ref{fig:egoexo_forecasting}, the error curves reveal a notable increase in prediction error over extended time frames. This trend aligns with expectations, considering the increasing unpredictability of human motion over time. The Oracle approach, serving as an upper bound for performance, evaluates solely the quality of pose prediction, excluding translational aspects. The considerable difference between  EgoCast and the Oracle suggests that, while EgoCast effectively predicts poses, further refinement is needed in correctly estimating the trajectory. When comparing the performance between EgoCast and the Location-based approach, we find that EgoCast consistently performs better across all timeframes. This finding is consistent with the improvements shown for current-frame pose estimation.  

Figure \ref{fig:ADT_results} illustrates the forecasting performance on the ADT dataset~\cite{pan2023aria}. Notably, the performance aligns with the results observed for the Ego-Exo4D dataset~\cite{grauman2023ego}, where our EgoCast approach significantly reduces the forecasting error compared to the previous state-of-the-art method.

Figure \ref{fig:qualitative} qualitatively shows the performance of our EgoCast baseline against the ground truth for a temporal horizon of \num{4} seconds on EgoExo-4D~\cite{grauman2023ego} and ADT~\cite{pan2023aria}. On the one hand, our method can predict accurate poses that resemble the ground truth. On the other hand, the predicted trajectories suffer from irregularities, indicating high translation changes between frames. Since forecasting in a large temporal window is a problem with several degrees of freedom, the method struggles to learn smooth transitions between frames. However, given the complex setup of predicting motion \num{4} seconds in the future, our method can accurately create plausible poses and trajectories.


Figure \ref{fig:egoexo_categories} presents the forecasting error for the eight activities within the Ego-Exo4D~\cite{grauman2023ego} dataset. Activities like music and health show minimal variations in both trajectory and pose, resulting in the lowest AUC values (\num{9.02}cm and \num{11.49}cm). In contrast, bike repairing and cooking involve limited pose adjustments but require large trajectory alterations due to spatial movements associated with these tasks (\eg going to the other side of the room to pick up a tool and returning to use it on a bike). Dance is characterized by minor trajectory shifts but significant pose variations. Although the requirements of the activity are the opposite of the cooking and bike repair, the error rates are similar. The most challenging activities for accurate forecasting are soccer and basketball, which demand frequent pose and trajectory changes reflected in the highest MPJPE values as the prediction time increases. These results highlight that complexity in translations and pose inherent to each activity varying over time significantly impacts the forecasting performance. Moreover, the per-category analysis of the ADT Database~\cite{pan2023aria}, presented in Figure 1 of the Supplementary Material, supports the ability of our EgoCast approach to predict plausible poses and trajectory movements. These findings underscore the applicability of our framework across different datasets for human pose estimation and trajectory forecasting in real-world scenarios.



\section{Conclusion}
\label{sec:formatting}
In this paper, we present EgoCast, a new experimental framework aimed at human pose forecasting from an egocentric perspective. Our approach introduces a novel method for combining visual and proprioceptive data. We also devise a current-frame estimation module to generate pseudo-groundtruths to avoid using groundtruth poses as input at inference time.  We believe that our work lays the groundwork for subsequent studies in human pose forecasting with egocentric inputs, potentially driving significant advancements in the field of computer vision and interactive technologies.

\newpage
\section*{Supplementary Material}
\section*{Ego-Exo4D}
\label{sec:egoexo_stats}

\subsection*{Dataset statistics}

Ego-Exo4D dataset~\cite{grauman2023ego} encompasses seven activities: Soccer, Basketball, Cooking, Dance, Bike Repair, Music and Health. Each set of activities provides a rich set of human motion patterns and poses, which allows one to understand 3D human motion in real-case scenarios, making Ego-Exo4D~\cite{grauman2023ego} a rich and realistic framework for 3D human pose forecasting. Table~\ref{tab:statistics} shows the statistics for the EgoExo4D dataset portion compatible with our EgoCast framework.

The license for using the Ego-Exo4D dataset can be found here \url{https://ego4d-data.org/pdfs/Ego-Exo4D-Model-License.pdf}.

\subsection*{Implementation details}
\label{sec:implementation}

For the current-frame estimation module, we use Adam optimizer with a learning rate of $1\times10^{-4}$ and a batch size of 24. We train this module for $2\times10^{5}$ iterations. Then, we perform a second-stage finetuning that incorporates the visual stream. We use EgoVLP \cite{lin2022egocentric} as a visual encoder and start from pretrained weights. Then, we finetune for an additional $5\times10^{4}$ iterations. Similarly, we train the forecasting branch for $3\times10^{4}$ iterations. The past time window \( k \) is set to 20 frames for both current-pose estimation and forecasting. For our experimental setup, we utilized the standard train/test splits as proposed by EgoExo-4D~\cite{grauman2023ego}. We train our model using PyTorch on 2 NVIDIA Quadro RTX 8000 GPUs over four hours.

\begin{table}[t]
\centering

\setlength{\tabcolsep}{12pt} 
\resizebox{0.48\textwidth}{!}{%
\begin{tabular}{lS[table-format=2.0]S[table-format=6.0]S[table-format=4.0]S[table-format=8.0]}
\toprule
\textbf{Activity} & \multicolumn{1}{l}{\textbf{Takes}} & \multicolumn{1}{l}{\textbf{Annotated frames}} & \multicolumn{1}{l}{\textbf{Seconds}} & \multicolumn{1}{l}{\textbf{Keypoints}} \\ \midrule
Soccer      & 160 & 596109 & 20170 & 9232969 \\
Basketball    & 612 & 1005790 & 34044 & 16744792 \\
Cooking    & 334 & 2307796  & 82230 & 32512567 \\
Dance & 581 & 790148 & 27720  & 13134050 \\
Bike Repair      & 275  & 1163466  & 42320 & 17558150 \\ 
Music     & 170 & 222179  & 10856 & 3186588 \\ 
Health      & 240  & 1284327  & 45515  & 15664947 \\ \midrule
Total      & 2372 & 7369815 & 262855 & 108034063 \\ \bottomrule
\end{tabular}%
}

\caption{\textbf{Ego-Exo4d~\cite{grauman2023ego} 3D human pose statistics.} Summary of the statistics for the annotated 3D human poses of the Ego-Exo4D~\cite{grauman2023ego} dataset. Overall, the dataset includes more than 100 million annotated keypoints distributed across 2372 takes.}

\vspace{8pt} 

\label{tab:statistics}
\end{table}

\subsection*{Architecture details}
\label{sec:architecture}
In our transformer-based forecasting module, we begin by applying a linear embedding map the 70 input features into a 256-dimensional space. Subsequently, the data is processed through a transformer composed of three self-attention layers, each with eight heads. To maintain a consistent output dimensionality, we use adaptive average pooling, ensuring an ending dimension size of 256, regardless of the input length. The process concludes with another linear embedding to obtain the final forecasting output.
\subsection*{Ablation experiments}\label{sec:ablation} 

Table \ref{tab:ablation} shows the performance of EgoCast for 1 second forecasting under different past window sizes. We find that optimizing the past window size has a notable impact, with a window size of 20 frames achieving the lowest error (19.66) in comparison to using fewer or more frames. Thus, 20 frames is the sweet spot for getting the most accurate predictions. These results suggest that having just the right amount of past information is key; insufficient data leads to inadequate forecasting, while excessive information potentially deteriorates the model's effectiveness, possibly due to the introduction of contradictory context.

\begin{table}[t]
\centering

\setlength{\tabcolsep}{12pt} 
\resizebox{0.47\textwidth}{!}{%
\begin{tabular}{@{}cccccc@{}}
\toprule
\textbf{Window size} & 5 & 10 & \textbf{20} & 40 & 80 \\ \midrule
\textbf{MPJPE}       & 29.34 &  26.57 & \textbf{19.66} &  32.80 &  33.11  \\ \bottomrule
\end{tabular}
}
\caption{\textbf{Forecasting ablation experiments.} We evaluate the impact of the past window size on forecasting 1 second in the future (30 frames). We achieved the best performance by using a window size of 20 frames.}
\vspace{8pt}
\label{tab:ablation}
\end{table}
\section*{Aria Digital Twin}\label{sec:adt_appendix}
We present results in the Aria Digital Twin (ADT) \cite{pan2023aria} dataset to showcase the versatility and generalization capacity of EgoCast.

\subsection*{Dataset description}

The Aria Digital Twin (ADT)~\cite{pan2023aria} dataset includes densely annotated sequences with ground-truth 3D body poses and the measurements collected by the Aria devices~\cite{somasundaram2023project}: egocentric video at $30$ frames per second, and position and rotation of the camera at every frame.

Table~\ref{tab:statisticsADT} shows the statistics for the ADT dataset portion compatible with our EgoCast framework.
Overall, the dataset contains \num{211824} frames with 3D human pose annotations of \num{21} joints distributed over 74 captures, with an average length of \num{2876} frames (\num{96}{s}) per capture. Furthermore, ADT includes diverse activities showcasing a vast variety of human motions; the most common activity is partying, followed by working, decorating, and having a meal. We divide the ADT dataset into a training and testing set for our experiments, using a 50:50 proportion. We ensure a uniform distribution of activities for both folds. ADT presents a high diversity of trajectories and motion of the camera wearer for each activity in the dataset. While for some activities, such as meal and work, trajectories are densely located in certain rooms of the scenario (\textit{e.g.}, kitchen and dinner room), other activities like decoration are distributed across the scene. Additionally, each set of activities provides a rich set of human motion patterns and poses, which allows one to understand 3D human motion in real-case scenarios, making the ADT dataset a rich and realistic framework for studying 3D human pose forecasting.

\begin{table}[t]
\centering

\resizebox{1\columnwidth}{!}{%
\begin{tabular}{@{}l@{\hspace{0.8em}}S[table-format=2.0]@{\hspace{0.8em}}S[table-format=6.0]@{\hspace{0.8em}}S[table-format=4.0]@{\hspace{0.8em}}S[table-format=8.0]@{}}
\toprule
\textbf{Activity} & \textbf{Captures} & \textbf{Annotated frames} & \textbf{Seconds} & \textbf{Keypoints} \\ \midrule
Party      & 40 & 118398 & 3947 & 2486358 \\
Work       & 15 & 41332  & 1378 & 867972 \\
Decoration & 10 & 27497  & 917  & 577437 \\
Meal       & 9  & 24597  & 820  & 516537 \\ \midrule
Total      & 74 & 211824 & 7061 & 4448304 \\ \bottomrule
\end{tabular}%
}

\caption{\textbf{ADT 3D human pose statistics.} Summary of the statistics for the annotated 3D human poses of the ADT dataset. Overall, the dataset includes more than 4 million annotated keypoints distributed across 74 captures.}
\label{tab:statisticsADT}
\end{table}

The licence for using the ADT dataset can be found here \url{https://www.projectaria.com/datasets/adt/license/}.

\subsection*{Ablation experiments}

\begin{table}[]
\centering
\begin{tabular}{c@{\hspace{30pt}}c@{\hspace{30pt}}c}
\hline
\textbf{Window size} & \textbf{Visual} & \textbf{MPJPE} \\ \hline
5                    & \ding{55}               & 9.76           \\
10                   & \ding{55}               & 9.57           \\
20                   & \ding{55}              & 9.27           \\
40                   & \ding{55}               & 9.36           \\ \hline
20                   & \ding{51}              & \textbf{7.43}  \\ \hline
\end{tabular}

\caption{\textbf{Ablation experiments for the current-frame estimation module.} We perform ablation experiments that include the rotation of the head-mounted device, the past window size, and the impact of using the visual cues. We achieved the best performance by using proprioception inputs (rotation and translation), a window size of 20 frames, and integrating visual streams for prediction.}
\label{tab:ablationADT}
\end{table}
Table \ref{tab:ablationADT} shows the performance of the current-frame estimation module in the EgoCast baseline. This analysis shows that optimizing the past window size has a notable impact, with a window size of 20 frames achieving the lowest error (9.27) without visual cues. The most substantial improvement is observed when integrating visual streams, which reduces the MPJPE to 7.43, underscoring the importance of combining proprioceptive inputs and visual data for enhanced pose estimation accuracy.

\begin{figure}[!tbp]
   \centering
   \begin{subfigure}[t]{\linewidth}
   \resizebox{\linewidth}{!}{%
   \begin{tikzpicture}[/pgfplots/width=1.6\linewidth, /pgfplots/height=0.92\linewidth]
      \begin{axis}[
                   ymin=0,ymax=70,xmin=0.5,xmax=5,
                   xlabel=Forecasting seconds,
                   ylabel=MPJPE,
                   xlabel shift={-2pt},
                   ylabel shift={-3pt},
                   font=\large,
                   axis equal image=false,
                   enlargelimits=false,
                   clip=true,
                   grid style=solid, grid=both,
                   major grid style={white!85!black},
                   minor grid style={white!95!black},
                   ytick={0,10,...,80},
                   yticklabels={0,10,20,30,40,50,60,70},
                   xtick={0,1,...,5},
                   xticklabels={0.5,1,2,3,4,5},
                   minor xtick={0,1,...,5},
                   minor ytick={0,1,...,70},
                   legend style={at={(0.05,0.95)},
                                 anchor=north west},
                   legend cell align={left}]
     
      \addplot+[p_green,solid,mark=*,mark options={fill=p_green, scale=1.1},ultra thick] table[x=seconds,y=decoration_ours]{images/category.txt};
      \addlegendentry{Decor (\num{35.89 }{cm})}
      \addplot+[p_blue,solid,mark=*,mark options={fill=p_blue, scale=1.1},ultra thick] table[x=seconds,y=meal_ours]{images/category.txt};
      \addlegendentry{Meal (\num{24.32 }{cm})}
      \addplot+[p_orange,solid,mark=*,mark options={fill=p_orange, scale=1.1},ultra thick] table[x=seconds,y=work_ours]{images/category.txt};
      \addlegendentry{Work (\num{21.86 }{cm})}
      \addplot+[p_red,solid,mark=*,mark options={fill=p_red, scale=1.1},ultra thick] table[x=seconds,y=party_ours]{images/category.txt};
      \addlegendentry{Party (\num{25.44 }{cm})}
      
      \end{axis}
   \end{tikzpicture}}
   \end{subfigure}
   \caption{\textbf{ADT Forecasting Performance by Category} For each activity in the ADT dataset, we show the performance of our method when forecasting \{0.5, 1, 2, 3, 4, and 5\} seconds into future. Note that since the graph shows MPJPE, lower curves represent better performance.}
   \label{fig:ADT_results_category}
\end{figure}
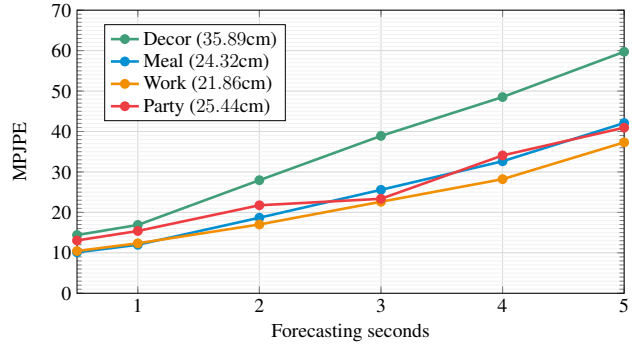

\subsection*{Results}

The per-category analysis in Figure \ref{fig:ADT_results_category} show that most activities within the ADT dataset have similar difficulty with the exception of decorating. The added difficulty in decorating comes from having large displacements throughout the environment. Note that our EgoCast approach is able to predict plausible poses and trajectory movements.

{\small
\bibliographystyle{ieee_fullname}
\bibliography{egbib}

\begin{thebibliography}{10}\itemsep=-1pt

\bibitem{akada20243d}
Hiroyasu Akada, Jian Wang, Vladislav Golyanik, and Christian Theobalt.
\newblock 3d human pose perception from egocentric stereo videos.
\newblock In {\em Proceedings of the IEEE/CVF Conference on Computer Vision and Pattern Recognition}, pages 767--776, 2024.

\bibitem{bouazizi2022motionmixer}
Arij Bouazizi, Adrian Holzbock, Ulrich Kressel, Klaus Dietmayer, and Vasileios Belagiannis.
\newblock Motionmixer: Mlp-based 3d human body pose forecasting.
\newblock In {\em Proceedings of the Thirty-First International Joint Conference on Artificial Intelligence, {IJCAI-22}}, pages 791--798. International Joint Conferences on Artificial Intelligence Organization, 7 2022.

\bibitem{briscoe2009egocentric}
Robert Briscoe.
\newblock Egocentric spatial representation in action and perception.
\newblock {\em Philosophy and Phenomenological Research}, 79(2):423--460, 2009.

\bibitem{castillo2023bodiffusion}
Angela Castillo, Maria Escobar, Guillaume Jeanneret, Albert Pumarola, Pablo Arbeláez, Ali Thabet, and Artsiom Sanakoyeu.
\newblock Bodiffusion: Diffusing sparse observations for full-body human motion synthesis.
\newblock In {\em Proceedings of the IEEE/CVF International Conference on Computer Vision}, 2023.

\bibitem{choudhury2023tempo}
Rohan Choudhury, Kris~M Kitani, and L{\'a}szl{\'o}~A Jeni.
\newblock Tempo: Efficient multi-view pose estimation, tracking, and forecasting.
\newblock In {\em Proceedings of the IEEE/CVF International Conference on Computer Vision}, pages 14750--14760, 2023.

\bibitem{Damen2018EPICKITCHENS}
Dima Damen, Hazel Doughty, Giovanni~Maria Farinella, Sanja Fidler, Antonino Furnari, Evangelos Kazakos, Davide Moltisanti, Jonathan Munro, Toby Perrett, Will Price, and Michael Wray.
\newblock Scaling egocentric vision: The epic-kitchens dataset.
\newblock In {\em European Conference on Computer Vision (ECCV)}, 2018.

\bibitem{Damen2021PAMI}
Dima Damen, Hazel Doughty, Giovanni~Maria Farinella, Sanja Fidler, Antonino Furnari, Evangelos Kazakos, Davide Moltisanti, Jonathan Munro, Toby Perrett, Will Price, and Michael Wray.
\newblock The epic-kitchens dataset: Collection, challenges and baselines.
\newblock {\em IEEE Transactions on Pattern Analysis and Machine Intelligence (TPAMI)}, 43(11):4125--4141, 2021.

\bibitem{Damen2022RESCALING}
Dima Damen, Hazel Doughty, Giovanni~Maria Farinella, Antonino Furnari, Jian Ma, Evangelos Kazakos, Davide Moltisanti, Jonathan Munro, Toby Perrett, Will Price, and Michael Wray.
\newblock Rescaling egocentric vision: Collection, pipeline and challenges for epic-kitchens-100.
\newblock {\em International Journal of Computer Vision (IJCV)}, 130:33–55, 2022.

\bibitem{du2023avatars}
Yuming Du, Robin Kips, Albert Pumarola, Sebastian Starke, Ali Thabet, and Artsiom Sanakoyeu.
\newblock Avatars grow legs: Generating smooth human motion from sparse tracking inputs with diffusion model.
\newblock In {\em Proceedings of the IEEE/CVF Conference on Computer Vision and Pattern Recognition}, pages 481--490, 2023.

\bibitem{fragkiadaki2015recurrent}
Katerina Fragkiadaki, Sergey Levine, Panna Felsen, and Jitendra Malik.
\newblock Recurrent network models for human dynamics.
\newblock In {\em Proceedings of the IEEE international conference on computer vision}, pages 4346--4354, 2015.

\bibitem{gong2023diffpose}
Jia Gong, Lin~Geng Foo, Zhipeng Fan, Qiuhong Ke, Hossein Rahmani, and Jun Liu.
\newblock Diffpose: Toward more reliable 3d pose estimation.
\newblock In {\em Proceedings of the IEEE/CVF Conference on Computer Vision and Pattern Recognition}, pages 13041--13051, 2023.

\bibitem{grauman2022ego4d}
Kristen Grauman, Andrew Westbury, Eugene Byrne, Zachary Chavis, Antonino Furnari, Rohit Girdhar, Jackson Hamburger, Hao Jiang, Miao Liu, Xingyu Liu, et~al.
\newblock Ego4d: Around the world in 3,000 hours of egocentric video.
\newblock In {\em Proceedings of the IEEE/CVF Conference on Computer Vision and Pattern Recognition}, pages 18995--19012, 2022.

\bibitem{grauman2023ego}
Kristen Grauman, Andrew Westbury, Lorenzo Torresani, Kris Kitani, Jitendra Malik, Triantafyllos Afouras, Kumar Ashutosh, Vijay Baiyya, Siddhant Bansal, Bikram Boote, et~al.
\newblock Ego-exo4d: Understanding skilled human activity from first-and third-person perspectives.
\newblock {\em arXiv preprint arXiv:2311.18259}, 2023.

\bibitem{guo2023back}
Wen Guo, Yuming Du, Xi Shen, Vincent Lepetit, Xavier Alameda-Pineda, and Francesc Moreno-Noguer.
\newblock Back to mlp: A simple baseline for human motion prediction.
\newblock In {\em Proceedings of the IEEE/CVF Winter Conference on Applications of Computer Vision}, pages 4809--4819, 2023.

\bibitem{huang2018deep}
Yinghao Huang, Manuel Kaufmann, Emre Aksan, Michael~J Black, Otmar Hilliges, and Gerard Pons-Moll.
\newblock Deep inertial poser: Learning to reconstruct human pose from sparse inertial measurements in real time.
\newblock {\em ACM Transactions on Graphics (TOG)}, 37(6):1--15, 2018.

\bibitem{jain2016structural}
Ashesh Jain, Amir~R Zamir, Silvio Savarese, and Ashutosh Saxena.
\newblock Structural-rnn: Deep learning on spatio-temporal graphs.
\newblock In {\em Proceedings of the ieee conference on computer vision and pattern recognition}, pages 5308--5317, 2016.

\bibitem{jiang2017seeing}
Hao Jiang and Kristen Grauman.
\newblock Seeing invisible poses: Estimating 3d body pose from egocentric video.
\newblock In {\em 2017 IEEE Conference on Computer Vision and Pattern Recognition (CVPR)}, pages 3501--3509. IEEE, 2017.

\bibitem{jiang2021egocentric}
Hao Jiang and Vamsi~Krishna Ithapu.
\newblock Egocentric pose estimation from human vision span.
\newblock In {\em 2021 IEEE/CVF International Conference on Computer Vision (ICCV)}, pages 10986--10994. IEEE, 2021.

\bibitem{jiang2022avatarposer}
Jiaxi Jiang, Paul Streli, Huajian Qiu, Andreas Fender, Larissa Laich, Patrick Snape, and Christian Holz.
\newblock Avatarposer: Articulated full-body pose tracking from sparse motion sensing.
\newblock In {\em Proceedings of European Conference on Computer Vision}. Springer, 2022.

\bibitem{lee2012discovering}
Yong~Jae Lee, Joydeep Ghosh, and Kristen Grauman.
\newblock Discovering important people and objects for egocentric video summarization.
\newblock In {\em 2012 IEEE conference on computer vision and pattern recognition}, pages 1346--1353. IEEE, 2012.

\bibitem{li2023ego}
Jiaman Li, Karen Liu, and Jiajun Wu.
\newblock Ego-body pose estimation via ego-head pose estimation.
\newblock In {\em Proceedings of the IEEE/CVF Conference on Computer Vision and Pattern Recognition}, pages 17142--17151, 2023.

\bibitem{li2018eye}
Yin Li, Miao Liu, and James~M Rehg.
\newblock In the eye of beholder: Joint learning of gaze and actions in first person video.
\newblock In {\em Proceedings of the European conference on computer vision (ECCV)}, pages 619--635, 2018.

\bibitem{lin2022egocentric}
Kevin~Qinghong Lin, Jinpeng Wang, Mattia Soldan, Michael Wray, Rui Yan, Eric~Z XU, Difei Gao, Rong-Cheng Tu, Wenzhe Zhao, Weijie Kong, et~al.
\newblock Egocentric video-language pretraining.
\newblock {\em Advances in Neural Information Processing Systems}, 35:7575--7586, 2022.

\bibitem{liu2022lower}
Mei Liu, Bo Peng, and Mingsheng Shang.
\newblock Lower limb movement intention recognition for rehabilitation robot aided with projected recurrent neural network.
\newblock {\em Complex \& Intelligent Systems}, 8(4):2813--2824, 2022.

\bibitem{liu2023egofish3d}
Yuxuan Liu, Jianxin Yang, Xiao Gu, Yijun Chen, Yao Guo, and Guang-Zhong Yang.
\newblock Egofish3d: Egocentric 3d pose estimation from a fisheye camera via self-supervised learning.
\newblock {\em IEEE Transactions on Multimedia}, 25:8880--8891, 2023.

\bibitem{luo2021dynamics}
Zhengyi Luo, Ryo Hachiuma, Ye Yuan, and Kris Kitani.
\newblock Dynamics-regulated kinematic policy for egocentric pose estimation.
\newblock {\em Advances in Neural Information Processing Systems}, 34:25019--25032, 2021.

\bibitem{luo2022embodied}
Zhengyi Luo, Shun Iwase, Ye Yuan, and Kris Kitani.
\newblock Embodied scene-aware human pose estimation.
\newblock {\em Advances in Neural Information Processing Systems}, 35:6815--6828, 2022.

\bibitem{martinez2017human}
Julieta Martinez, Michael~J Black, and Javier Romero.
\newblock On human motion prediction using recurrent neural networks.
\newblock In {\em Proceedings of the IEEE conference on computer vision and pattern recognition}, pages 2891--2900, 2017.

\bibitem{natsakis2021predicting}
Tassos Natsakis and Lucian Busoniu.
\newblock Predicting intention of motion during rehabilitation tasks of the upper-extremity.
\newblock In {\em 2021 43rd Annual International Conference of the IEEE Engineering in Medicine \& Biology Society (EMBC)}, pages 6037--6040. IEEE, 2021.

\bibitem{ng2020you2me}
Evonne Ng, Donglai Xiang, Hanbyul Joo, and Kristen Grauman.
\newblock You2me: Inferring body pose in egocentric video via first and second person interactions.
\newblock In {\em Proceedings of the IEEE/CVF Conference on Computer Vision and Pattern Recognition}, pages 9890--9900, 2020.

\bibitem{pan2023aria}
Xiaqing Pan, Nicholas Charron, Yongqian Yang, Scott Peters, Thomas Whelan, Chen Kong, Omkar Parkhi, Richard Newcombe, and Yuheng~Carl Ren.
\newblock Aria digital twin: A new benchmark dataset for egocentric 3d machine perception.
\newblock In {\em Proceedings of the IEEE/CVF International Conference on Computer Vision}, pages 20133--20143, 2023.

\bibitem{park2016egocentric}
Hyun~Soo Park, Jyh-Jing Hwang, Yedong Niu, and Jianbo Shi.
\newblock Egocentric future localization.
\newblock In {\em Proceedings of the IEEE Conference on Computer Vision and Pattern Recognition}, pages 4697--4705, 2016.

\bibitem{piumsomboon2018mini}
Thammathip Piumsomboon, Gun~A Lee, Jonathon~D Hart, Barrett Ens, Robert~W Lindeman, Bruce~H Thomas, and Mark Billinghurst.
\newblock Mini-me: An adaptive avatar for mixed reality remote collaboration.
\newblock In {\em Proceedings of the 2018 CHI conference on human factors in computing systems}, pages 1--13, 2018.

\bibitem{qiu2022egocentric}
Jianing Qiu, Lipeng Chen, Xiao Gu, Frank P-W Lo, Ya-Yen Tsai, Jiankai Sun, Jiaqi Liu, and Benny Lo.
\newblock Egocentric human trajectory forecasting with a wearable camera and multi-modal fusion.
\newblock {\em IEEE Robotics and Automation Letters}, 7(4):8799--8806, 2022.

\bibitem{somasundaram2023project}
Kiran Somasundaram, Jing Dong, Huixuan Tang, Julian Straub, Mingfei Yan, Michael Goesele, Jakob~Julian Engel, Renzo De~Nardi, and Richard Newcombe.
\newblock Project aria: A new tool for egocentric multi-modal ai research.
\newblock {\em arXiv preprint arXiv:2308.13561}, 2023.

\bibitem{su2023review}
Dongnan Su, Zhigang Hu, Jipeng Wu, Peng Shang, and Zhaohui Luo.
\newblock Review of adaptive control for stroke lower limb exoskeleton rehabilitation robot based on motion intention recognition.
\newblock {\em Frontiers in Neurorobotics}, 17, 2023.

\bibitem{su2016detecting}
Yu-Chuan Su and Kristen Grauman.
\newblock Detecting engagement in egocentric video.
\newblock In {\em Computer Vision--ECCV 2016: 14th European Conference, Amsterdam, The Netherlands, October 11-14, 2016, Proceedings, Part V 14}, pages 454--471. Springer, 2016.

\bibitem{tang20233d}
Zhenhua Tang, Zhaofan Qiu, Yanbin Hao, Richang Hong, and Ting Yao.
\newblock 3d human pose estimation with spatio-temporal criss-cross attention.
\newblock In {\em Proceedings of the IEEE/CVF Conference on Computer Vision and Pattern Recognition}, pages 4790--4799, 2023.

\bibitem{tome2019xr}
Denis Tome, Patrick Peluse, Lourdes Agapito, and Hernan Badino.
\newblock xr-egopose: Egocentric 3d human pose from an hmd camera.
\newblock In {\em Proceedings of the IEEE/CVF International Conference on Computer Vision}, pages 7728--7738, 2019.

\bibitem{wang2024egocentric}
Jian Wang, Zhe Cao, Diogo Luvizon, Lingjie Liu, Kripasindhu Sarkar, Danhang Tang, Thabo Beeler, and Christian Theobalt.
\newblock Egocentric whole-body motion capture with fisheyevit and diffusion-based motion refinement.
\newblock In {\em Proceedings of the IEEE/CVF Conference on Computer Vision and Pattern Recognition}, pages 777--787, 2024.

\bibitem{wang2021estimating}
Jian Wang, Lingjie Liu, Weipeng Xu, Kripasindhu Sarkar, and Christian Theobalt.
\newblock Estimating egocentric 3d human pose in global space.
\newblock In {\em Proceedings of the IEEE/CVF International Conference on Computer Vision}, pages 11500--11509, 2021.

\bibitem{wang2023scene}
Jian Wang, Diogo Luvizon, Weipeng Xu, Lingjie Liu, Kripasindhu Sarkar, and Christian Theobalt.
\newblock Scene-aware egocentric 3d human pose estimation.
\newblock In {\em Proceedings of the IEEE/CVF Conference on Computer Vision and Pattern Recognition}, pages 13031--13040, 2023.

\bibitem{wang2021deep}
Jinbao Wang, Shujie Tan, Xiantong Zhen, Shuo Xu, Feng Zheng, Zhenyu He, and Ling Shao.
\newblock Deep 3d human pose estimation: A review.
\newblock {\em Computer Vision and Image Understanding}, 210:103225, 2021.

\bibitem{wang2023holoassist}
Xin Wang, Taein Kwon, Mahdi Rad, Bowen Pan, Ishani Chakraborty, Sean Andrist, Dan Bohus, Ashley Feniello, Bugra Tekin, Felipe~Vieira Frujeri, et~al.
\newblock Holoassist: an egocentric human interaction dataset for interactive ai assistants in the real world.
\newblock In {\em Proceedings of the IEEE/CVF International Conference on Computer Vision}, pages 20270--20281, 2023.

\bibitem{xu2023eqmotion}
Chenxin Xu, Robby~T Tan, Yuhong Tan, Siheng Chen, Yu~Guang Wang, Xinchao Wang, and Yanfeng Wang.
\newblock Eqmotion: Equivariant multi-agent motion prediction with invariant interaction reasoning.
\newblock In {\em Proceedings of the IEEE/CVF Conference on Computer Vision and Pattern Recognition}, pages 1410--1420, 2023.

\bibitem{yan2024forecasting}
Haitao Yan, Qiongjie Cui, Jiexin Xie, and Shijie Guo.
\newblock Forecasting of 3d whole-body human poses with grasping objects.
\newblock In {\em Proceedings of the IEEE/CVF Conference on Computer Vision and Pattern Recognition}, pages 1726--1736, 2024.

\bibitem{yi2022physical}
Xinyu Yi, Yuxiao Zhou, Marc Habermann, Soshi Shimada, Vladislav Golyanik, Christian Theobalt, and Feng Xu.
\newblock Physical inertial poser (pip): Physics-aware real-time human motion tracking from sparse inertial sensors.
\newblock In {\em Proceedings of the IEEE/CVF Conference on Computer Vision and Pattern Recognition}, pages 13167--13178, 2022.

\bibitem{yi2021transpose}
Xinyu Yi, Yuxiao Zhou, and Feng Xu.
\newblock Transpose: Real-time 3d human translation and pose estimation with six inertial sensors.
\newblock {\em ACM Transactions on Graphics (TOG)}, 40(4):1--13, 2021.

\bibitem{yuan2019ego}
Ye Yuan and Kris Kitani.
\newblock Ego-pose estimation and forecasting as real-time pd control.
\newblock In {\em Proceedings of the IEEE/CVF International Conference on Computer Vision}, pages 10082--10092, 2019.

\bibitem{zhang2023probabilistic}
Siwei Zhang, Qianli Ma, Yan Zhang, Sadegh Aliakbarian, Darren Cosker, and Siyu Tang.
\newblock Probabilistic human mesh recovery in 3d scenes from egocentric views.
\newblock {\em arXiv preprint arXiv:2304.06024}, 2023.

\bibitem{zhang2022egobody}
Siwei Zhang, Qianli Ma, Yan Zhang, Zhiyin Qian, Taein Kwon, Marc Pollefeys, Federica Bogo, and Siyu Tang.
\newblock Egobody: Human body shape and motion of interacting people from head-mounted devices.
\newblock In {\em European Conference on Computer Vision}, pages 180--200. Springer, 2022.

\bibitem{zhang2024dynamic}
Yu Zhang, Songpengcheng Xia, Lei Chu, Jiarui Yang, Qi Wu, and Ling Pei.
\newblock Dynamic inertial poser (dynaip): Part-based motion dynamics learning for enhanced human pose estimation with sparse inertial sensors.
\newblock In {\em Proceedings of the IEEE/CVF Conference on Computer Vision and Pattern Recognition}, pages 1889--1899, 2024.

\bibitem{zheng2022gimo}
Yang Zheng, Yanchao Yang, Kaichun Mo, Jiaman Li, Tao Yu, Yebin Liu, C~Karen Liu, and Leonidas~J Guibas.
\newblock Gimo: Gaze-informed human motion prediction in context.
\newblock In {\em European Conference on Computer Vision}, pages 676--694. Springer, 2022.

\end{thebibliography}
}

\end{document}